\documentclass[10pt,letterpaper]{article}

\usepackage{cogsci}

\usepackage{pslatex}
\usepackage{array,multirow}
\usepackage{apacite}
\usepackage{graphicx}
\usepackage{lineno,hyperref}
\usepackage{amsmath} 
\usepackage{natbib}
\usepackage{eurosym}
\usepackage{graphicx}
\usepackage{wrapfig}
\usepackage[english]{babel}
\usepackage[center]{caption}
\usepackage{mathtools}
\usepackage{mathptmx}
\usepackage{latexsym}
\usepackage{url}
\usepackage{multirow}
\usepackage{amssymb}
\usepackage{capt-of, blindtext}
\usepackage{pdfpages}
\usepackage{setspace}
\usepackage{lineno}
\usepackage{algorithmic}
\usepackage{algorithm}
\usepackage{fullpage}
\usepackage{kpfonts}
\usepackage{setspace}
\usepackage{algorithmic}
\usepackage{algorithm}
\usepackage{bbold}

\title{QuLBIT: Quantum-Like Bayesian Inference Technologies for Cognition and Decision}
 
\author{
        {\large \bf Catarina Moreira$^1$ (catarina.pintomoreira@qut.edu.au)} \AND 
        {\large \bf Matheus Hammes$^1$ (matheus.hammes@connect.qut.edu.au)} \AND
        {\large \bf Rasim Serdar Kurdoglu$^2$ (r.s.kurdoglu@bilkent.edu.tr)} \AND
        {\large \bf Peter Bruza$^1$ (p.bruza@qut.edu.au)} \AND
        $^1$School of Information Systems, Queensland University of Technology, Brisbane, Australia\\
        $^2$Faculty of Business Administration, Bilkent University, Ankara, Turkey
}

\begin{document}

\maketitle

\begin{abstract} 

This paper provides the foundations of a unified cognitive decision-making framework (QulBIT) which is derived from quantum theory. The main advantage of this framework is that it can cater for paradoxical and irrational human decision making. Although quantum approaches for cognition have demonstrated advantages over classical probabilistic approaches and bounded rationality models, they still lack explanatory power. To address this, we introduce a novel explanatory analysis of the decision-maker's belief space.
This is achieved by exploiting quantum interference effects as a way of both quantifying and explaining the decision-maker’s uncertainty. We detail the  main  modules  of the unified framework, the explanatory analysis method, and illustrate their application in situations violating the Sure Thing Principle.


\end{abstract}

\subsection{Introduction}

The primary motivation behind QuLBIT\footnote{\url{https://github.com/catarina-moreira/QuLBiT}} is the challenge to formally account for seemingly paradoxical human decision making. It is widely known in the literature of cognitive science and economics that when it comes to decision-making under uncertainty, humans usually make decisions that are inconsistent with the axioms of expected utility theory, leading to decisions that are either sub-optimal, paradoxical or even irrational~\citep{Kahneman79,Moreira17lestah}. These paradoxical decisions can result in cognitive biases~\citep{Tversky74}, violations of major economic principles (like the Sure Thing Principle)~\citep{savage54,Allais53,Ellsberg61} or violations to the laws of probability theory and logic~\citep{busemeyer06,Busemeyer09}. In short, decades of research has found a whole range of human judgements that deviates substantially from what would be considered normatively correct according to logic or probability theory. 

\begin{figure*}[ht!]
    \centering
    \includegraphics[scale=0.143]{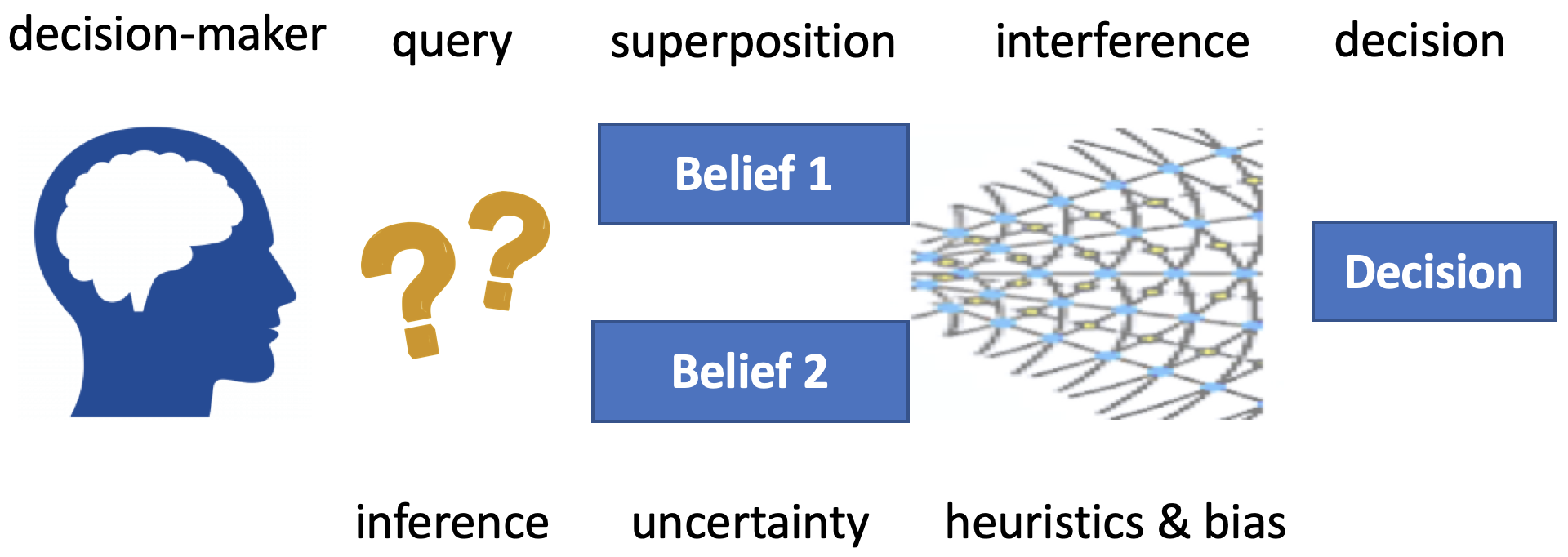}
    \caption{The representation of the decision-maker's beliefs under uncertainty where beliefs can enter into a superposition and suffer non-linear quantum interference effects that can destroy or reinforce certain beliefs in a single time step, leading to decisions that deviate from classical theory.}
    \label{fig:general_cog_persp}
\end{figure*}

 The field of Quantum Cognition emerged to respond to this challenge, a major feature being the use of quantum probability theory to model human cognition, including decision making~\citep{Busemeyer12book}.
Quantum probability can be viewed as generalisation of  Bayesian probability theory. 
In quantum-like cognitive models, events are modelled as sub-spaces of a Hilbert spaces, a vector space of complex numbers (amplitudes) which enables the calculation of probabilities by projection: performing the squared magnitude of an amplitude. This representation allows events to interfere with each other, which influences their associated probabilities. These interference effects generate a set of new parameters that can be used to either accommodate violations in Bayesian theory~\citep{Busemeyer12book} or paradoxical human decisions~\cite{Kahneman82book}.  Interference is a core concept in the QuLBIT framework which enables an alternative quantification of uncertainty, as well as the representation of conflicting, ambiguous beliefs. The vector representation of superposition obeys neither the distributive axiom of Boolean logic nor the law of total probability~\citep{Moreira16review} (See Figure~\ref{fig:general_cog_persp}). 

Quantum cognitive models make use of additional parameters, which allows fitting to empirical data but which ``do not necessarily explain them" \citep{Blutner16}.  QuLBIT addresses this lack of explanatory power by employing a novel analysis method which allows interpretation of the decision-maker's belief space, e.g., when a decision-maker prefers one choice over another.
This paper illustrates QuLBIT's novel explanatory analysis method in regard to violations of the Sure Thing Principle in the Prisoner's Dilemma Game~\citep{Shafir92}.



 

\section{The Prisoner's Dilemma Game and Violations to the Sure Thing Principle}

Several experiments in the literature have shown that people violate this principle in decisions under uncertainty, leading to paradoxical results and violations of the law of
total probability~\citep{Tversky74,Aerts04,Birnbaum08,Li02,Hristova08}. The prisoner's dilemma is an example where, under uncertainty, people violate the sure thing principle, by being more cooperative.


To test the veracity of the Sure Thing Principle under the Prisoner’s Dilemma game, experiments were made in where three conditions were tested:
\begin{itemize}
    \item Participants were informed the other participant chose to
\textit{Defect} (\textbf{Condition 1:} Known to defect);
    \item Participants were informed the other participant chose to
\textit{Cooperate} (\textbf{Condition 2:} Known to cooperate);
    \item Participants had \textit{no information} about the other participant’s
decision (\textbf{Condition 3 :} Unknown).
\end{itemize}

\begin{table}[h!]
\resizebox{\columnwidth}{!} {
    \begin{tabular}{| l | c  |}
    \hline
    \textbf{Condition}  & \textbf{Pr( P2 = Defect)}    \\
    \hline
    Condition 1 (P1 Known to Defect):            &  0.97                  \\
    \hline
     Condition 2 (P1 Known to Cooperate) :           &  0.84    \\
    \hline
     Condition 3:  (P1 Unknown)            &  0.63                         \\
    \hline
    \end{tabular}
    }
    \caption{Experimental results from \citet{Shafir92} PD experiment. 
    }
    \label{tab:summary_lit}
\end{table}

Table~\ref{tab:summary_lit} summarizes the results of these experiments for the three conditions. The column classical prediction shows the classical probability of a player choosing to $Defect$, given that the decision of Player 1 is unknown (Condition 3). The payoff matrix used in \citet{Shafir92} Prisoner's Dilemma experiment can be found in Table~\ref{tab:util}.

\begin{table}[h!]
\centering
\begin{tabular}{c c | c}
            & P2 = Def  & P2 = Coop \\
            \cline{2-3}
 P1 = Def   & 30        & 25         \\
 P1 = Coop  & 85        & 36           \\
            \cline{2-3}
\end{tabular}
\caption{Payoff matrix used in \citet{Shafir92} Prisoner's Dilemma experiment.}
\label{tab:util}
\end{table}
 


\section{QuLBIT: Quantum-Like Bayesian Inference Technologies}

The QuLBIT framework provides a unifying decision model for cognitive decision making, which is susceptible to cognitive biases.
In addition, the framework caters for data-driven computational decisions,
which are based on optimization algorithms. The  main  advantage of this framework is that it can cater for paradoxical and irrational human decisions during the
inference process.  
This not only enhances the understanding of cognitive decision-making, but is also relevant for providing effective decision support systems.   The nature of the quantum-like approach allows the system to capture optimal, sub-optimal (bounded rational), or even irrational decisions which play an important role in a ``humanistic system", which are systems strongly influenced by  human judgment, and behaviour.

\subsection{Views on Rationality}

\begin{figure}[h!]
\centering
    \includegraphics[scale=0.09]{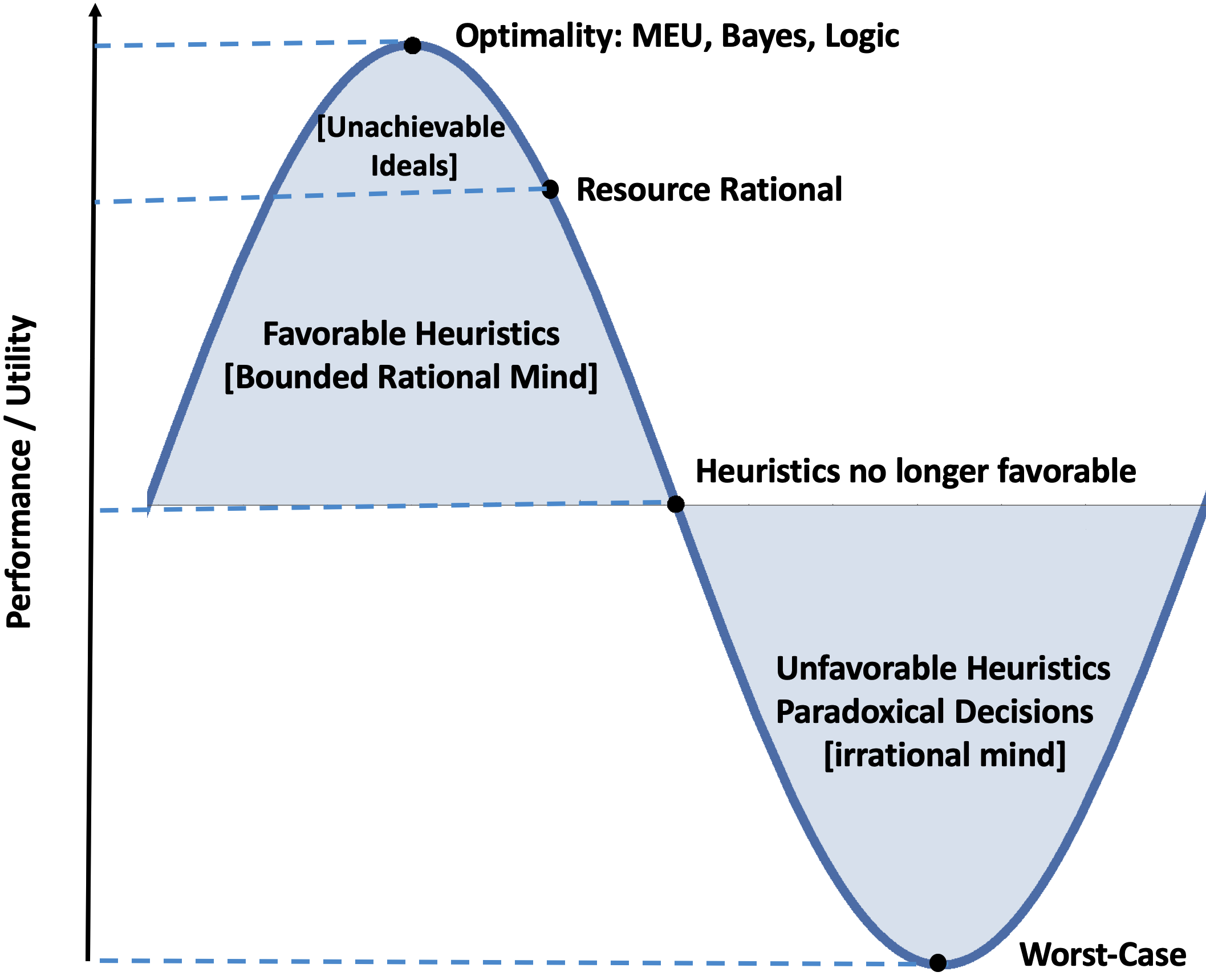}
    \caption{Wave-like interpretation of the QuLBIT framework in terms of the notions of rationality.}
    \label{fig:views}
\end{figure}

The QuLBIT framework caters for a spectrum of views in relation to rational decision making, depending on the degrees of rationality that the decision-maker employs. 
The views presented under the proposed quantum-like approach are similar to the ones put forward by~\citet{Gigerenzer96} and~\citet{Lieder19bbs}. They include decisions bounded in terms of time, processing power, information, etc. We extend this view to incorporate the notion of the \textit{irrational mind}, a point in the decision-maker's belief space where heuristics are no longer sufficient to produce satisficing outcomes. Figure~\ref{fig:views} illustrates the different views on rationality that are represented in the QuLBIT framework depicted in terms of a non-linear quantum interference wave. We define these views in the following way: 

\begin{itemize}

\item \textbf{Belief space: } Corresponds to the set of possible beliefs that are held by a decision-maker from the moment that (s)he is faced with a decision until (s)he actually makes the decision. Beliefs as inputs of thought, and desires as motivational sources of reasoning, guide decision-making~\citep{Cushman19}. 

\item \textbf{Unbounded Rationality: } Corresponds to the ideal scenario where the decision-maker has unlimited cognitive resources:  time, processing power and information in order to transact the decision. Note that the ideal scenario is often not within reach for a human decision-maker.

\item \textbf{Bounded Rationality: } Corresponds to the scenario where the decision-maker makes decisions bounded in terms of cognitive resources with limited information, time, and processing power. Consequently, fast and frugal heuristics are applied sometimes resulting in sub-optimal decisions~\citep{Kahneman82book}, but also yielding good-enough adaptive decisions~\citep{Gigerenzer11, Spiliopoulos20}, i.e. favourable in terms of meeting desires satisfactorily without creating illogical or improbable conclusions. Fast and frugal heuristics, by their definition, are not produced by strictly logical or probabilistic calculations, but equally they do not violate logical or probabilistic principles. 

\item \textbf{Resource Rationality: } Corresponds to bounded rationality in the decision-maker's belief space that lead to the maximum performance for a given level of uncertainty~\citet{Lieder19bbs}.

\item \textbf{Irrationality: } Corresponds to the situation where bounded rationality can no longer apply fast and frugal heuristics that satisfy the desires due to extreme levels of uncertainty or due to the inability of the decision-maker to make adequate decisions. These decisions generally have poor utility and performances. The fast and frugal heuristics start to fail, and the decision-maker experiences cognitive dissonance effects and starts to violate rules of logic and probability, which in turn leads to unfavourable heuristics and their paradoxical and irrational outcomes. These decisions are hard (or even impossible) to be captured by current computational decision systems. 

\end{itemize}

\subsection{Quantum-Like Bayesian Networks}

\begin{figure}[h!]
   \centering
    \includegraphics[scale=0.27]{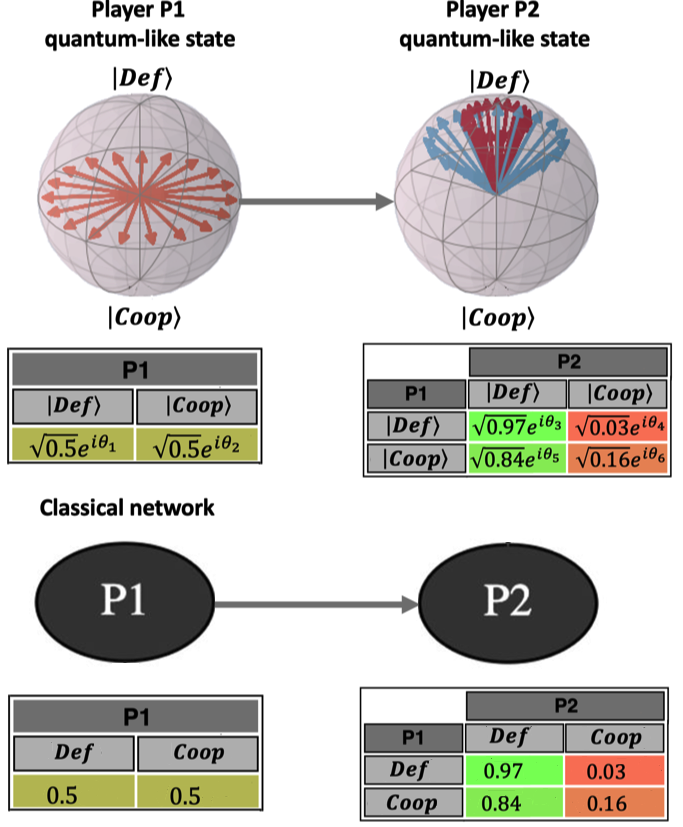}
    \caption{Representation of \citet{Shafir92} PD experiment using a quantum-like BN (left) and a classical BN.}
    \label{fig:qlbn}
\end{figure}

The fundamental core of the QuLBIT framework is the notion of graphical probabilistic inference using the formalism of quantum theory. The Quantum-like Bayesian network, originally proposed in~\citet{Moreira14,Moreira15,Moreira16,Moreira20bbs} is the fundamental building block of the QuLBIT framework and is also an example of a model that has been extensively studies in the literature for predicting and accommodating paradoxical human decisions across different decision scenarios, ranging from psychological experiments~\citep{Moreira16,Moreira17faces,Wichert20} to real-world credit application scenarios~\citep{Moreira18bpmn}. The difference between the traditional Bayesian network (BN) and a quantum-like Bayesian Network (QLBN) is the way one specifies the values of the conditional probability tables. While in the traditional BNs, one uses real numbers to express probabilities, in quantum mechanics these probabilities are expressed as probability amplitudes, which are represented by complex numbers. Figure~\ref{fig:qlbn} shows a representation of a Quantum-Like Bayesian Network.


Exact inference in QLBNs is given by three steps:

\begin{itemize}
    \item \textbf{Definition of the superposition state.} 
    In quantum theory, all individual quantum states contained in a Hilbert Space are defined by a superposition state which is represented by a quantum state vector $| S \rangle$ comprising the occurrence of all events of the system. This can be analogous to the classical full joint probability distribution, with the difference that instead of using real numbers to express probabilities, one uses complex probability amplitudes. For example, for \citet{Shafir92} PD experiment, the full joint distribution and the corresponding superposition state can be represented by the following vectors:
    \[
    Joint = \left[
    \begin{matrix}
         ~~0.485~~ \\
         ~~0.015~~ \\
         ~~0.420~~ \\
         ~~0.080~~ \\
    \end{matrix}
    \right]
    ~~~~~S = \left[
    \begin{matrix}
        \sqrt{0.485} ~ e^{i \theta_1}~~ \\
        \sqrt{0.015} ~ e^{i \theta_2}~~ \\
        \sqrt{0.420} ~ e^{i \theta_3}~~ \\
        \sqrt{0.080} ~ e^{i \theta_4}~~ \\
    \end{matrix}
    \right]
    \]

    \item \textbf{A density matrix which describes the quantum system}.
    The density operator, $\rho$, aims to describe a system where we can compute the probabilities of finding each state in the network. 
    One way to achieve this is by computing a density operator through the product between the superposition state $S$ and the corresponding conjugate transpose $S^\dagger$, that is $\rho = S S^\dagger$.
  
    \begin{figure}[h!]
        \resizebox{\columnwidth}{!} {
        \includegraphics{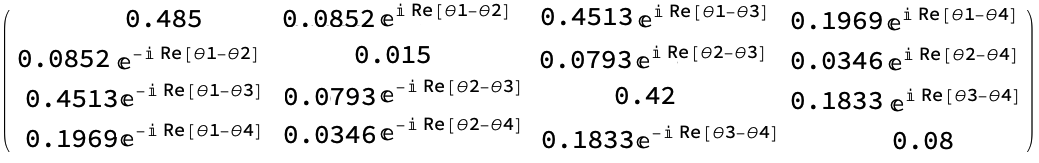}
        }
        \caption{Density matrix, $\rho =  S S^\dagger $}
        \label{fig:my_label}
    \end{figure}
    
    The density operator also contains quantum interference terms in the off-diagonal elements, which are the core of this model. It is precisely through these off-diagonal elements that, during the inference process, one is able to obtain quantum interference effects, and consequently, deviations from normative probabilistic inferences. 

    \item \textbf{Quantum-Like Marginal Distribution}. 
    The quantum-like marginal probability can be defined by two selection operators $Def$ and $Coop$, which are vectors that select the entries of the classical joint distribution that match the query.

\[ Def = \left[
    \begin{matrix}
    1 \\
    0 \\
    1 \\
    0 \\
    \end{matrix}
    \right]
    ~~~Coop = \left[
    \begin{matrix}
    0 \\
    1 \\
    0 \\
    1 \\
    \end{matrix}
    \right]
\]
    Computing the quantum-like probability of Player 2 defecting, given that (s)he is uncertain about Player 1's strategy, $Pr_q(P2 = Def)$, corresponds to summing out all entries of the joint related to Player 1's strategy. This probability can be computed by applying the selection operator to the density matrix, $\rho$, and normalising the results with a normalisation factor $\gamma = 1/(Pr_q( P2 = Def ) + Pr_q( P2 = Coop )$, and where $Def^T$ is the transpose of operator $Def$,
    \begin{equation}
        \begin{split}
            Pr_q(P2 = Def) = \gamma| Def \rho Def^T | = \\
             ~~~~~~~~ = \gamma|  0.905 + 2 \times 0.451331 \cos( \theta_{Def} ) 
        \end{split}
    \end{equation}

    This is the same as having the classical probability, $Pr( P2 = Def )$, together with a quantum interference term, $Interf_{Def}$, which corresponds to the emergence of destructive / constructive interference effects, associated with the uncertainty that the player is experiencing,
    \begin{equation}
        \begin{split}
        Pr_q(P2 = Def) = \gamma| Pr(P2 = Def ) + 2 Interf_{Def} | = \\
        \gamma| 0.905 + 2 \times 0.451331 \cos( \theta_{Def} ) | 
    \end{split}
    \end{equation}
    In the same way, we can compute the probability of Player 2 cooperating, $Pr_q( P2 = Coop)$,
    \begin{equation}
        \begin{split}
             Pr_q(P2 = Coop) = \gamma| Coop \rho Coop^T | = \\
             ~~~~~~~~ = \gamma|  0.095  + 2 \times 0.034641 \cos( \theta_{Coop} )
        \end{split}
    \end{equation}
    This suggests that the proposed model provides a hierarchy of mental representations ranging from quantum-like effects to pure classical ones. 

\end{itemize}

Note that if one sets $\theta_{Def}$ or $\theta_{Coop}$ to $\pi /2$, then $\cos (\theta_{Def})=0$ and $\cos (\theta_{Coop})=0$. This means that the interference term is canceled and the quantum-like Bayesian network collapses to its classical counterpart. 
Setting $\theta_{Def} = \theta_{Coop} = 2.8057$, will reproduce the disjuction effect observed in~\citet{Shafir92}, $Pr_q( P2 = Def ) = 0.63$. In \citet{Moreira16,Moreira20Balanced}, the authors proposed a similarity heuristic and later a law of balance that are able to automatically find the values of $\theta_{Def}$ and $\theta_{Coop}$ without manually fitting the data. 

\subsection{Quantum-Like Influence Diagrams}

Quantum-Like Influence diagrams \citep{Moreira18qi} are a directed acyclic graph structure that represents a full probabilistic description of a decision problem by using probabilistic inferences performed in quantum-like Bayesian networks together with an utility function.


 Given a set of possible decision rules, $\delta_A$, a classical Influence Diagram computes the decision rule that leads to the Maximum Expected Utility in relation to decision $D$. In a classical setting, this formula makes use of a full joint probability distribution, $Pr_{\delta_A}( x | a )$, over all possible outcomes, $x$, given different actions $a$ belonging to the decision rules $\delta_A$ where the goal is to choose some action $a$ that maximises the expected utility with respect to some decision rule, $\delta_A$.

\begin{equation}
EU\left[ \mathcal{D}\left[\delta_A \right] \right] = \sum_{x, a} Pr{\delta_A}\left( x, a \right) U \left( x , a \right).
\label{eq:EU_init}
\end{equation}

The quantum-like approach of the influence diagrams consists in replacing the classical probability, $Pr{\delta_A}$, by its quantum counterpart, $Pr_q{\delta_A}$.
The general idea is to take advantage of the quantum interference terms produced in the quantum-like Bayesian network to influence the probabilities used to compute the expected utility. 

Mathematically, one can define utility operators, $U_{Def}$ and $U_{Coop}$, that represent the payoff that Player 2 receives if (s)he chooses to $Defect$ and to $Cooperate$, respectively. And the quantum-like influence diagram simply consists in replacing the classical probability $Pr_{\delta_A}( x, a )$ in Equation \ref{eq:EU_init}, by the probability computed by the quantum-like Bayesian network $Pr_q{\delta_D}( x | a )$. Details of these formalisms can be found in the publicly available notebook\footnote{\url{https://git.io/JfKKB}} and \citet{Moreira18qi}.

\[ U_{Def} = \left[
    \begin{matrix}
    30  & 0 & 0 & 0 \\
    0  & 0 & 0 & 0 \\
    0  & 0 & 85 & 0 \\
    0  & 0 & 0 & 0 \\
    \end{matrix}
    \right]
    U_{Coop} = \left[
    \begin{matrix}
    0  & 0 & 0 & 0 \\
    0  & 25 & 0 & 0 \\
    0  & 0 & 0 & 0 \\
    0  & 0 & 0 & 36 \\
    \end{matrix}
    \right]
\]

The expected utility of Player 2 defecting becomes
\begin{equation}
    EU_q\left[ \mathcal{D}\left[\delta_{Def} \right] \right] =\sum_{x, def} Pr_{\delta_{Def}}\left( x, def \right) U \left( x , def \right)
\end{equation}
\begin{equation}
     EU_q\left[ P2 = Def \right] = Trace\left[ Pr_q(P2 = Def) U_{Def} \right] 
\end{equation}
\begin{equation}
 EU_q\left[ P2 = Def \right] = 
\frac{115 \cos (\theta_{Def}) + 115.02}{0.131 \cos (\theta_{Coop} + \cos (\theta_{Def}) + 1.132}
\label{eq:util_def}
\end{equation}
In the same way, the expected utility of Player 2 cooperating becomes
\begin{equation}
\begin{split}
EU_q\left[ P2 = Coop \right] = Trace\left[ Pr_q(P2 = Coop) U_{Coop} \right] \\ 
= \frac{100 \cos ( \theta_{Coop}) + 137.121}{ \cos (\theta_{Coop})+13.0288 \cos (\theta_{Def}) +14.4338}
\end{split}
\label{eq:util_coop}
\end{equation}

From this formalism, the region of the belief space where the decision-maker will always perceive that (s)he will have a higher utility for cooperating, $EU_q\left[ P2 = Coop \right] = X > X = EU_q\left[ P2 = Def \right]$, is given by  $\theta_{Def} = \pi$, and $ 0 \geq \theta_{Coop} \leq 2 \pi$.

\section{A Novel Explanatory Analysis in Quantum-Like Decision Models}

\begin{figure*}[h!]
    \resizebox{2\columnwidth}{!} {
    \includegraphics{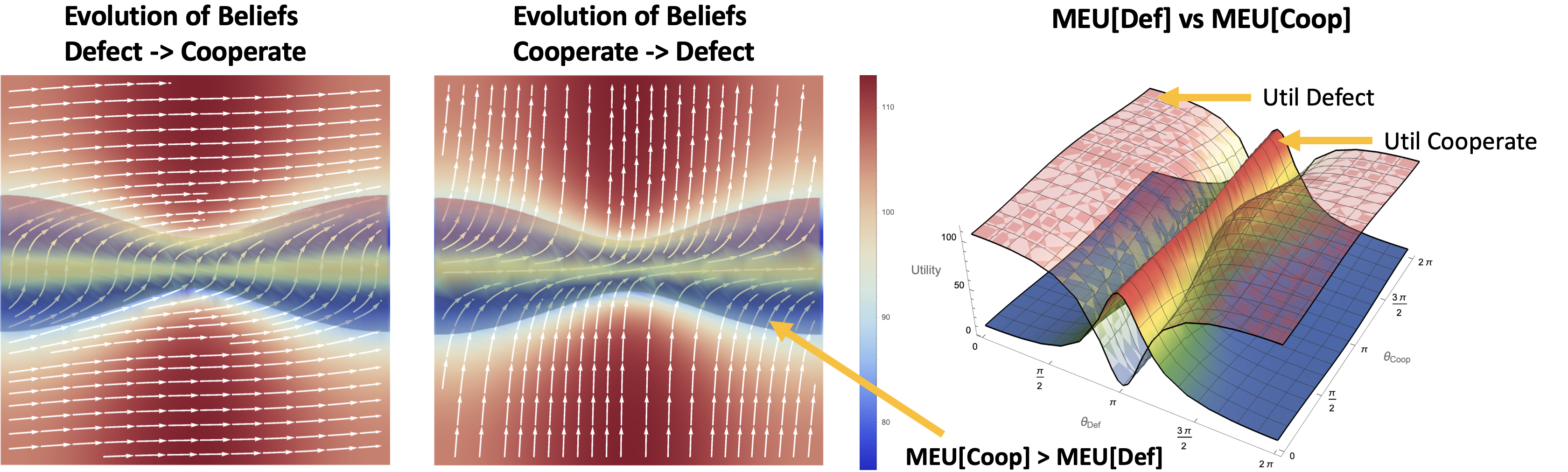}
    }
    \caption{Evolution of the quantum interference waves in the decision-maker's belief space for ~\citet{Shafir92} PD experiment.}
    \label{fig:evolution}
\end{figure*}

In the QuLBIT framework, for decisions under uncertainty, the decision-maker's beliefs are represented as waves during the reasoning process. Only when the decision-maker makes a decision, these beliefs collapse to the chosen decision with a certain probability and utility. Before reaching a decision, the decision-maker can experience uncertainty regarding Player 1's actions. This corresponds to beliefs of $cooperate$ and $defect$ competing with each other causing constructive/destructive interferences (quantum interference parameters $\theta_{Def}$ and $\theta_{Coop}$). 

 Figure~\ref{fig:evolution} (right) shows the combined graphical representations of the utilities that a player can obtain when reasoning about considering a $Defect$ strategy (Equations~\ref{eq:util_def}) or a $Cooperate$ strategy (Equation~\ref{eq:util_coop}) according to the uncertainty that (s)he feels about Player 1's actions. 
 

It follows from Figure~\ref{fig:evolution} (right) that this model allows different levels of representations of decisions under uncertainty ranging from (1) fully rational and optimal decisions ({\it fully classical}), (2) sub-optimal decisions, 
to (3) irrational decisions~\citep{Shafir92} ({\it fully quantum}). Figures~\ref{fig:evolution} (left) and Figures~\ref{fig:evolution} (center) show the evolution of the decision-maker's beliefs through the belief space enabling a novel analysis and interpretation of the quantum interference waves in terms of the different levels of rationality.  


\begin{itemize}
\item {\bf Fully classical decisions:} the majority of the decision-makers are stable in the regions of the belief space where the MEU of defect is maximised.  This notion is in accordance with predictions from expected utility theory and concepts from Game Theory, where in strictly dominant strategies, the decision-maker stays stable in the Nash equilibrium state, in this case, engaging in a defect strategy. In this region, it seems that the decision-maker is not experiencing much uncertainty, and consequently quantum interference effects are minimum.

\item {\bf Sub-Optimal decisions:} the lighter regions of the figure indicate decisions where the MEU of defecting is close to the MEU of cooperate, $MEU(Defect) \approx MEU( Cooperate )$, but still the decision-maker prefers to defect, because there are not enough heuristic cues to convince him/her to switch from defect to cooperate. 
Quantum interference effects occur and uncertainty is high, however the quantum interference effects are not strong enough to make the decision-maker change his mind and for that reason (s)he continues to choose according to expected utility. 

\item {\bf Irrational decisions:} correspond to the central, blue regions of the Figure~\ref{fig:evolution}. In these regions, uncertainty is maximised and quantum interference effects are significant enough to make the decision-maker change his mind. The decision-maker irrationally engages in wishful thinking, or beliefs that are far stretched from available data on hand.
It is in this region where the decision-maker perceives that $MEU(Cooperate) > MEU(Defect)$, and consequently makes a decision that deviates from the classical notions of the Expected Utility theory. Notions of game theory actually accept the fact that the decision-maker might not always obey to the formalisms of expected utility theory. This  can occur when players did not understand the rules of the game, or simply because they played randomly. What game theory notions tell us is that the decision-maker will not stay stable in these irrational states (in this case the $Cooperate$ state), which is in accordance with Figure~\ref{fig:evolution}. 
\end{itemize}

\section{Conclusions}

It is the purpose of this paper to provide a set of contributions of quantum-based models applied to cognition and decision as an alternative mathematical approach for decision-making in order to better understand the structure of human behaviour.

The QuLBIT framework is open-source and allows the combination of both Bayesian and non-Bayesian influences in cognition, where classical representations provide a better account of data as individuals gain familiarity, and quantum-like representations can provide novel predictions and novel insights about the relevant psychological principles involved during the decision process.

Our contributions so far show that QuLBIT is a unified framework for cognition and decision-making that is able (1) to accommodate and predict paradoxical human decisions (namely disjunction errors), (2) to analyse the belief space of the decision-maker through quantum interference, (3) to quantify uncertainty and (4) to provide a non-linear view on the different levels of rationality, ranging from fully optimal decisions (classical) to irrational decisions (quantum-like).

\bibliographystyle{apacite}

\setlength{\bibleftmargin}{.125in}
\setlength{\bibindent}{-\bibleftmargin}


\end{document}